\documentclass[10pt,twocolumn,letterpaper]{article}

\usepackage{iccv}
\usepackage{times}
\usepackage{epsfig}
\usepackage{graphicx}
\usepackage{amsmath}
\usepackage{amssymb}
\usepackage{booktabs}
\usepackage{multirow}

\usepackage{url}

\usepackage[pagebackref=true,breaklinks=true,letterpaper=true,colorlinks,bookmarks=false]{hyperref}

\iccvfinalcopy 

\def\iccvPaperID{560} 

\ificcvfinal\fi
\begin{document}

\title{Differential Recurrent Neural Networks for Action Recognition}

\author{Vivek Veeriah\\
University of Central Florida\\
4000 Central Flordia Blvd\\
{\tt\small vivekveeriah@knights.ucf.edu}
\and
Naifan Zhuang\\
University of Central Florida\\
4000 Central Flordia Blvd\\
{\tt\small zhuangnaifan@knights.ucf.edu}
\and
Guo-Jun Qi\thanks{Corresponding author}\\
University of Central Florida\\
4000 Central Flordia Blvd\\
{\tt\small
guojun.qi@ucf.edu}
}

\maketitle

\begin{abstract}
The long short-term memory (LSTM) neural network is capable of processing complex sequential information since it utilizes special gating schemes for learning representations from long input sequences. It has the potential to model any sequential time-series data, where the current hidden state has to be considered in the context of the past hidden states. This property makes LSTM an ideal choice to learn the complex dynamics of various actions. Unfortunately, the conventional LSTMs do not consider the impact of spatio-temporal dynamics corresponding to the given salient motion patterns, when they gate the information that ought to be memorized through time. To address this problem, we propose a differential gating scheme for the LSTM neural network, which emphasizes on the change in information gain caused by the salient motions between the successive frames. This change in information gain is quantified by Derivative of States (DoS), and thus the proposed LSTM model is termed as differential Recurrent Neural Network (dRNN). We demonstrate the effectiveness of the proposed model by automatically recognizing actions from the real-world 2D and 3D human action datasets. Our study is one of the first works towards demonstrating the potential of learning complex time-series representations via high-order derivatives of states.
\end{abstract}

\section{Introduction}
Recently, Recurrent Neural Networks (RNNs) \cite{schuster1997bidirectional}, especially Long Short-Term Memory (LSTM) model \cite{hochreiter1997long}, have gained significant attention in solving many challenging problems involving sequential time-series data, such as action recognition \cite{gregor2015draw,donahue2014long,grushin2013robust}, multilingual machine translation \cite{sutskever2014sequence,bahdanau2014neural}, multimodal translation between videos and sentences \cite{venugopalan2014translating}, and robot control \cite{mayer2008system}.
In these applications, learning an appropriate representation of sequences is an important step in achieving artificial intelligence.

Compared with many existing spatio-temporal features \cite{klaser2008spatio,scovanner20073} from the time-series data, RNN use either a hidden layer \cite{schuster1997bidirectional} or a memory cell \cite{hochreiter1997long} to learn the time-evolving states which models the underlying dynamics of the input sequence.
For example, \cite{baccouche2010action,donahue2014long} have used LSTM to model the video sequences to learn their long short-term dynamics.  In contrast to the conventional RNN, the major component of LSTM is the memory cell which is modulated by three gates - input, output and forget gates. These gates determine the amount of dynamic information entering/leaving the memory cell. The memory cell has a set of internal states, which store the information obtained over time.  In this context, these internal states constitute a representation of an input sequence learned over time.

In many recent works, the LSTMs have shown tremendous potential in action recognition tasks \cite{baccouche2010action}\cite{grushin2013robust}\cite{donahue2014long}.
The existing LSTM model represents a video by integrating over time all the available information from each frame.  However, we observed that for an action recognition task, not all frames contain salient spatio-temporal information which are discriminative to different classes of actions. Many frames contain non-salient motions which are irrelevant to the performed action.

This inspired us to develop a new family of LSTM model that automatically learns the dynamic saliency of the actions performed. The conventional LSTM fails to learn the salient dynamic patterns comprehensively, since the gate units do not explicitly consider whether a frame contains salient motion information when they modulate the memory cells. Thus the model is insensitive to the dynamic evolution of the hidden states given the input video sequences. To address this problem, we propose the differential RNN (dRNN) model that learns these salient spatio-temporal representations of actions.

Specifically, dRNN models the dynamics of actions by computing different-orders of Derivative of States (DoS) that are sensitive to the spatio-temporal structure of actions.  Depending on the DoS, the gate units can learn the appropriate information that should be required to model the dynamic evolution of actions. To train the dRNN model, we use truncated Back Propagation algorithm to prevent the exploding or diminishing errors through time \cite{hochreiter1997long}. In particular, we follow the rule that the errors propagated through the connections to those DoS nodes would be truncated once they leave the current memory cell.


Finally, we demonstrate that the dRNNs can achieve the state-of-the-art performance on both 2D and 3D action recognition datasets. Specifically, dRNNs outperform the existing LSTM model on these action recognition tasks, consistently achieving the better performance with the same input sequences.  On the other hand, when compared with the other algorithms tailored to model special assumptions on spatio-temporal structure of actions, the proposed general-purpose dRNN model can still reach competitive performance.

The remainder of this paper is organized as follows.  In the next section \ref{sec:rel}, we review several related work to the action recognition problem. The background and details of RNNs and LSTMs are reviewed in section \ref{sec:bg}.  Section \ref{sec:drnn} presents the proposed differential RNNs (dRNNs). The experimental results are presented in section \ref{sec:exp}. Finally, we conclude and discuss the future work related to dRNNs in section \ref{sec:clu}.

\section{Related Work}\label{sec:rel}
Action recognition has been a long-standing research problem in computer vision and pattern recognition community, which aims to enable a computer to automatically understand the activities performed by people interacting with the surrounding environment and with each other \cite{poppe2010survey}. This is a challenging problem due to the huge intra-class variance of actions performed by different actors at various speeds, in diverse environments (e.g., camera angles, lighting conditions, and cluttered background).

To address this problem, many robust spatio-temporal representations have been constructed. For example, HOG3D \cite{klaser2008spatio} uses the histogram of 3D gradient orientations to represent the motion structure over the frame sequences; 3D-SIFT \cite{scovanner20073} extends the popular SIFT descriptor to characterize the scale-invariant spatio-temporal structure for 3D video volume; actionlet ensemble \cite{wang2012mining} utilizes a robust approach to model the discriminative features from 3D positions of the tracked joints captured by depth cameras.

Although these descriptors have achieved remarkable success, they are usually engineered to model a specific spatio-temporal structure in an ad-hoc fashion. Recently, the huge success of deep networks in image classification \cite{krizhevsky2012imagenet} and speech recognition \cite{graves2014towards} has inspired many researchers to apply the deep neural networks, such as 3D Convolutional Neural Networks (3DCNNs) \cite{baccouche2011sequential} and Recurrent Neural Networks (RNNs) \cite{baccouche2010action,donahue2014long}, to action recognition.  In particular, \cite{baccouche2011sequential} developed a 3D convolutional neural network that extends the conventional CNN by taking space-time volume as input. On the contrary, \cite{baccouche2010action,donahue2014long} used LSTMs to represent the video sequences directly, and modeled the dynamic evolution of the action states via a sequence of memory cells.

Meanwhile, the existing approaches combine deep neural networks with spatio-temporal descriptors, achieving competitive performance. For example, in \cite{baccouche2011sequential}, a LSTM model takes a sequence of Harris3D and 3DCNN descriptors extracted from each frame as input, and the result on KTH dataset has shown the state-of-the-art performance \cite{baccouche2011sequential}.

\section{Background}\label{sec:bg}
In this section, we briefly review the recurrent neural network as well as its variant, long short-term memory model.  Readers who are familiar with them might skip to the next section directly.

\subsection{Recurrent Neural Networks}
Traditional recurrent neural networks (RNNs) \cite{schuster1997bidirectional} model the dynamics of an input sequence of frames $\{\mathbf x_t\in\mathbb R^n| t=1,\cdots,T\}$ through
a sequence of hidden states $\{\mathbf s_t\in\mathbb R^m|t=1,\cdots,T\}$ thereby learning the spatio-temporal structure of the input sequence.  For example, a classical RNN model uses the following recurrent equation
\begin{equation}
\mathbf s_{t}=\tanh( \mathbf W_{ss}\mathbf s_{t-1} + \mathbf W_{sx}\mathbf x_{t} + \mathbf b_s)
\end{equation}
to model the hidden state $\mathbf s_{t}$ at time $t$ by combining the information from the current input $\mathbf x_t$ and the past hidden state $\mathbf s_{t-1}$, where the hyperbolic tangent $\tanh(\cdot)$ is an activation function with range $[-1,1]$, $\mathbf W_{sx}$ and $\mathbf W_{ss}$ are two mapping matrices to the hidden state, and $\mathbf b_s$ is the bias vector.

The hidden state can be mapped to an output sequence $\{\mathbf z_t\in\mathbb R^k|t=1,\cdots,T\}$ as
\begin{equation}
\mathbf z_{t}=\tanh(\mathbf W_{zs} \mathbf s_{t} + \mathbf b_{z})
\end{equation}
where each $\mathbf z_{t}$ represents an $1$-of-$k$ encoding of the confidence scores on $k$ classes of actions.  Then, this output vector can be transformed to a vector of probabilities $\mathbf y_{t}$ by softmax function as
\begin{equation}\label{Eq:softmax}
y_{t,c}=\dfrac{\exp(z_{t,c})}{\sum\limits_{l=1}^{k}\exp(z_{t,l})},
\end{equation}
with each entry $y_{t,c}$ being the probability of frame $t$ belonging to class $c\in\{1,\cdots,k\}$.

\subsection{Long Short-Term Memory}
The above classical RNN is limited in learning the long-term representation of video sequences, due to the exponential decay in retaining the context information of video frames \cite{hochreiter1997long}.  To overcome this limitation, Long Short-Term Memory (LSTM) \cite{hochreiter1997long}, a variant of RNN, has been designed to learn the long-range dependency between the output label and the input frame, which has achieved competitive performance on action recognition task \cite{baccouche2010action}\cite{baccouche2011sequential}.

In particular, LSTMs are composed of a sequence of memory cells, each containing
 an internal state $\mathbf s_t$ storing the memory of the input sequence up to time $t$.
To store the memory with respect to a context in long period of time, three types of gate units are incorporated into LSTMs to control what information would enter and leave the memory cell over time \cite{hochreiter1997long}.  These gate units are activated by a nonlinear function of input/output sequences as well as internal states, making them powerful enough to model the dynamically changing context given that the human actions evolve at various time scales.

Formally, a LSTM cell has the following gates:

{\noindent\bf 1. The input gate} $\mathbf i_{t}$ controls the degree to which the input information would enter the memory cell to influence its internal state $\mathbf s_{t}$ at time $t$.  The activation of this gate has the following recurrent form
      \[
      \mathbf i_{t}=\sigma(\mathbf W_{is} \mathbf s_{t-1} + \mathbf W_{iz}  \mathbf z_{t-1} + \mathbf W_{ix}  \mathbf x_{t} + \mathbf b_i)
      \]
      where the sigmoid $\sigma(\cdot)$ is an activation function with the range $[0,1]$, with $0$ meaning the gate is closed and $1$ meaning the gate is completely open; $\mathbf W_{i*}$ are the mapping matrices and $\mathbf b_i$ is the bias vector.


{\noindent\bf 2. The forget gate} $\mathbf f_{t}$ modulates the previous state $\mathbf s_{t-1}$  to control its contribution to the current state (c.f. Eq(\ref{Eq:update})).  It is defined as
      \[
      \mathbf f_{t}=\sigma(\mathbf W_{fs} \mathbf s_{t-1} + \mathbf W_{fz} \mathbf z_{t-1} + \mathbf W_{fx} \mathbf x_{t} + \mathbf b_f)
      \]
      with the mapping matrices $\mathbf W_{f*}$ and the bias vector $\mathbf b_f$.

  With the input/forget gate units, the internal state $\mathbf s_{t}$ of each memory cell can be updated below:
\begin{equation}\label{Eq:update}
\mathbf s_{t}=\mathbf f_{t}\odot \mathbf s_{t-1} + \mathbf i_{t}\odot \mathbf s_{t-\frac{1}{2}}
\end{equation}
where we define the pre-state $\mathbf s_{t-\frac{1}{2}}$ as
\[
\mathbf s_{t-\frac{1}{2}}= \tanh(\mathbf W_{sz}\mathbf z_{t-1} + \mathbf W_{sx}\mathbf x_{t}+\mathbf b_s).
\]
The pre-state can be considered as an intermediate state between two consecutive frames, aggregating the information from the last output $\mathbf z_{t-1}$ and the current input $\mathbf x_{t}$.  Then it is combined with the gated information from the previous state $\mathbf s_{t-1}$ to update the current state $\mathbf s_{t}$ as in Eq.~(\ref{Eq:update}).


{\noindent\bf 3. The output gate} $\mathbf o_{t}$:
\[
\mathbf o_{t}=\sigma(\mathbf W_{os} \mathbf s_{t} + \mathbf W_{oz} \mathbf z_{t-1} + \mathbf W_{ox} \mathbf x_{t} + \mathbf b_o).
\]
It gates the information output from a memory cell which would influence the future states of LSTM cells. Then the output of a memory cell can be expressed as
\begin{equation}\label{Eq:zoutput}
\mathbf z_{t}=\mathbf o_{t}\odot \tanh(\mathbf W_{zs} \mathbf s_{t} + \mathbf b_{z})
\end{equation}
where $\odot$ stands for element-wise product.

In brief, LSTM proceeds by iteratively applying Eq.~(\ref{Eq:update}) and Eq.~(\ref{Eq:zoutput}) to update the state $\mathbf s_t$ and output $\mathbf z_t$.  In this process, the forget gate, output gate and input gate play a critical role in
controlling the information entering and leaving the memory cell. More details about LSTMs can be found in \cite{hochreiter1997long}.

\section{Differential Recurrent Neural Networks}\label{sec:drnn}

For an action recognition task, not all video frames contain salient patterns to discriminate between different classes of actions. Many spatio-temporal descriptors, such as  3D-SIFT \cite{scovanner20073} and HoGHoF \cite{laptev2008learning}, have been proposed to localize and encode the salient spatio-temporal points. They detect and encode the spatio-temporal points related to salient motions of the objects in video frames, revealing the important dynamics of actions.

In this paper, we develop a novel LSTM model to automatically learn the dynamics of actions, by detecting and integrating the salient spatio-temporal sequences. The conventional LSTMs might fail to capture these salient dynamic patterns, because the gate units
do not {\em explicitly} consider the impact of dynamic structures present in input sequences.  This makes the model inadequate to learn the evolution of action states. To address this problem, we propose a differential RNN (dRNN) model to learn and integrate the dynamics of actions.

\begin{figure}[t]
\begin{center}
   \includegraphics[width=0.9\linewidth]{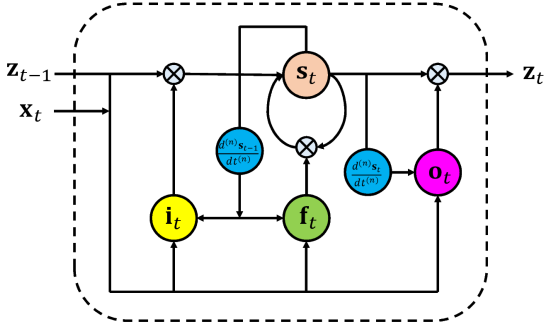}
\end{center}
   \caption{Architecture of the proposed dRNN model at time $t$. In the memory cell, the input gate $\mathbf i_t$ and the forget gate $\mathbf f_t$ are controlled by DoS $\frac{d^{(n)}\mathbf s_{t-1}}{dt^{(n)}}$ at $t-1$, and the output gate $\mathbf o_t$ is controlled by the DoS $\frac{d^{(n)}\mathbf s_{t}}{dt^{(n)}}$ at $t$.}
\label{Fig:dRNN}
\end{figure}

The proposed dRNN model is based on the observation that the internal state of each memory cell contains the accumulated information about the spatio-temporal structure, i.e., it is a long short-term representation of an input sequence. So the Derivative of States (DoS) $\dfrac{d\mathbf s_t}{dt}$
quantifies the change of information at each time $t$.
In other words, a large magnitude of DoS is an indicator of a salient spatio-temporal structure containing the informative dynamics caused by an abrupt change of action state.  In this case, the gate units should allow more information to enter the memory cell to update its internal state. Otherwise, when the magnitude of DoS is small, the incoming information should be gated out of the memory cell so the internal state would not be affected by the current input.  Therefore, DoS should be used as a factor to gate the information flow into and out of the internal state of memory cell over time.

Moreover, we can involve higher-orders of DoS $\{\dfrac{d^n \mathbf s_t}{dt^n}|n\geq 2\}$ to detect and capture the higher-order dynamic patterns for the dRNN model. For example, when modeling a moving object in a video,  the first-order DoS captures the velocity while the second-order captures its acceleration.  These different orders of DoS will enable dRNN to better represent the dynamic evolution of action states.


Figure \ref{Fig:dRNN} illustrate the architecture of the proposed dRNN model. Formally, we have the following recurrent equations to control the gate units with the DoS up to order $N$:
\begin{equation}\label{Eq:input}
\small
\mathbf i_{t}=\sigma(\sum_{n=0}^N \mathbf W_{id}^{(n)} \frac{d^{(n)}\mathbf s_{t-1}}{dt^{(n)}} +\mathbf W_{iz} \mathbf z_{t-1} + \mathbf W_{ix} \mathbf x_{t} + \mathbf b_{i})
\end{equation}
\begin{equation}\label{Eq:forget}
\small
\mathbf f_{t}=\sigma( \sum_{n=0}^N \mathbf W_{fd}^{(n)}\dfrac{d^{(n)}\mathbf s_{t-1}}{dt^{(n)}} + \mathbf W_{fz} \mathbf z_{t-1} + \mathbf W_{fx} \mathbf x_{t} + \mathbf b_{f})
\end{equation}
\begin{equation}\label{Eq:output}
\small
\mathbf o_{t}=\sigma( \sum_{n=0}^N \mathbf W_{od}^{(n)}\dfrac{d^{(n)}\mathbf s_{t}}{dt^{(n)}} + \mathbf W_{oz} \mathbf z_{t-1} + \mathbf W_{ox} \mathbf x_{t} + \mathbf b_{o})
\end{equation}
where $\dfrac{d^{(n)}\mathbf s_{t-1}}{dt^{(n)}}$ is the $n$-order DoS, and
$W_{*d}^{(n)}$ are the corresponding mapping matrices.


Finally, it is worth pointing out that we do not use the derivative of inputs as a measurement of salient dynamics to control the gate units. The derivative of inputs would amplify the unwanted noises which are often contained in the input sequence. This derivative of inputs only represent the local dynamic saliency, in contrast to the long short-term change in the information gained over time. For example, a motion may have been performed several frames ago. Using derivative of inputs would treat it as a novel salient motion, even though it has already been stored by LSTM. On the contrary, DoS does not have this problem, because the internal state $\mathbf s_t$ has long-term memory of the past motion pattern, even though the same motion had previously occurred.

\subsection{Discretized Model}
Since the model is defined in the discrete-time domain, the first-order derivative $\dfrac{d\mathbf s_t}{dt}$, as the velocity of information change, can be
discretized as the difference of states
\begin{equation}\label{Eq:v}
\mathbf v_{t}\triangleq\dfrac{d\mathbf s_t}{dt} \doteq \mathbf s_{t}-\mathbf s_{t-1}
\end{equation}
for its simplicity \cite{epperson2013numerical}.

Similarly, we can consider the second order of DoS as the acceleration of information change can be discretized as
\begin{equation}\label{Eq:a}
\mathbf a_{t}\triangleq\dfrac{d^2\mathbf s_t}{dt^2}\doteq \mathbf v_{t}-\mathbf v_{t-1}=\mathbf s_{t}-2\mathbf s_{t-1}+\mathbf s_{t-2}
\end{equation}
In this paper, we only consider the first two orders of DoS.  Higher orders can be derived in a similar way.
With the above recurrent equations, at time step $t$, the dRNN model proceeds in the following order.
\begin{itemize}
  \item Compute input gate activation $\mathbf i_t$ and forget gate activation $\mathbf f_t$ by Eq.~(\ref{Eq:input}) and Eq.~(\ref{Eq:forget});
  \item Update state $\mathbf s_t$ with $\mathbf i_t$ and $\mathbf f_t$ by Eq.~(\ref{Eq:update});
  \item Compute discretized DoS $\{\dfrac{d^{(n)} \mathbf s_{t}}{dt^{(n)}}|n=1,\cdots,N\}$ up to order $N$ at time $t$, e.g. Eq.~(\ref{Eq:v}) and Eq.~(\ref{Eq:a});
  \item Compute output gate $\mathbf o_t$ by Eq.~(\ref{Eq:output});
  \item Output $\mathbf z_t$ gated by $\mathbf o_t$ from memory cell by Eq.~(\ref{Eq:zoutput});
  \item (Optional) Output the label $\mathbf y_{t}$ by applying the softmax to $\mathbf z_t$ by Eq.~(\ref{Eq:softmax}).
\end{itemize}



Now it is obvious that this model is termed differential RNNs (dRNNs) because of the central role of  derivatives of states in detecting and capturing the salient spatio-temporal structures.

\subsection{Learning Algorithm}

To learn the model parameters of dRNNs, we define a loss function to measure the deviation between the target class $c_t$ and $\mathbf y_t$ at time $t$:
\[
\ell(\mathbf y_t, c_t) = -\log y_{t,c_t}.
\]

For an action recognition task, the label of action is often given at the video level. Since LSTMs have the ability to memorize the content of an entire sequence, the last memory cell of LSTMs ought to contain all the necessary information for action recognition. Thus, for a sequence of length $T$, and a given training label $c$, the dRNNs can be trained by minimizing the loss at time $T$, i.e., $\ell(\mathbf y_T, c) = -\log y_{T,c}$.

Otherwise, if an individual label $c_t$ is given to each frame $t$ in the sequence, we can minimize the cumulative loss over the sequence:
$$
\sum_{t=1}^T {\ell(\mathbf y_t, c_t)}.
$$

Both types of loss functions can be minimized by Back Propagation Through Time (BPTT) \cite{cuellar2006application}, which unfolds a dRNN model over several time steps and then runs the back propagation algorithm to train the model.
To prevent the back-propagated errors from decaying or exploding exponentially, LSTMs usually use truncated BPTT \cite{hochreiter1997long}. The idea is rather simple: once the back-propagated error leaves the memory cell or gates, it will not be allowed to enter the memory cell again.  In the proposed dRNNs, we also use the truncated errors to learn the model parameters.  In particular, we do not allow the errors to re-enter the memory cell once they leave it through the DoS nodes $\mathbf v_{t}$ and $\mathbf a_{t}$.


Formally, we assume the following truncated derivatives of gate activations:
$$
\dfrac{\partial\mathbf i_t}{\partial \mathbf v_{t-1}} \circeq \mathbf 0,~~\dfrac{\partial\mathbf f_t}{\partial \mathbf v_{t-1}} \circeq \mathbf 0,~~\dfrac{\partial\mathbf o_t}{\partial \mathbf v_t} \circeq \mathbf 0
$$
and
$$
\dfrac{\partial\mathbf i_t}{\partial \mathbf a_{t-1}} \circeq \mathbf 0,~~\dfrac{\partial\mathbf f_t}{\partial \mathbf a_{t-1}} \circeq \mathbf 0,~~\dfrac{\partial\mathbf o_t}{\partial \mathbf a_t} \circeq \mathbf 0
$$
where $\circeq$ stands for the truncated derivative.  The details about the implementation of truncated BPTT can be found in \cite{hochreiter1997long}.



\section{Experiments and Results}\label{sec:exp}
We compare the performance of the proposed method with the state-of-the-art LSTM and non-LSTM methods present in existing literature on both 2D and 3D human action datasets.

\subsection{Datasets}
The proposed method is evaluated on the KTH 2D action recognition dataset, as well as MSR Action3D dataset.

{\bf KTH dataset.} We choose KTH dataset \cite{schuldt2004recognizing} for it is a {\em de facto} benchmark for evaluating action recognition algorithms. This makes it possible to directly compare with the other algorithms. There are two KTH datasets - KTH-1 and KTH-2, which both consist of six action classes: walking, jogging, running, boxing, hand-waving and hand-clapping. The actions are performed several times by $25$ subjects in four different scenarios: outdoors, outdoors with scale variation, outdoors with different clothes and indoors. The sequences are captured over homogeneous background with a static camera recording $25$ frames per second. Each video has a resolution of $160 \times 120$, and lasts for about 4 seconds on KTH-1 dataset and for about a second for KTH-2 dataset. There are $599$ videos in the KTH-1 dataset and $2,391$ video sequences in the KTH-2 dataset.

{\bf MSR Action3D dataset.}  The MSR Action3D dataset \cite{li2010action} consists of $567$ depth map sequences performed by $10$ subjects using a depth sensor similar to the Kinect device. The resolution of each video is $320 \times 240$ and there are $20$ action classes where each subject performs each action two or three times. The actions are chosen in the context of gaming. They cover a variety of movements related to arms, legs, torso etc. This dataset has a lot of noise in the joint locations of the skeleton as well as high intra-class variations and inter-class similarities, making it a challenging dataset for evaluation among the existing 3D datasets. We follow a similar experiment setting from \cite{wang2012mining}, where half of the subjects are used for training and the other half are used for testing. This setting is much more challenging than the subset one used in \cite{li2010action}, because all actions are evaluated together and the chance of confusion is much higher.

\subsection{Feature Extraction}
We are using densely sampled HOG3D features to represent each frame of video sequences from the KTH dataset.
Specifically, we uniformly divide the 3D video volumes into a dense grid, and extract the descriptors from each cell of the grid.
The parameters for HOG3D are the same as the one used in \cite{klaser2008spatio}. We extract HOG3D features using the standard KTH optimized dense sampling parameters mentioned on the authors' webpage \footnote{\url{http://lear.inrialpes.fr/people/klaeser/software_3d_video_descriptor}}. The size of the descriptor was 1000 per cell of grid, and there are $56$ such cells in each frame, yielding a $56,000$ dimensional feature vector per frame. We apply PCA to reduce the dimension to $450$, retaining 97\% of energy among the principal components, to construct a compact input into the dRNN model.

For 3D action dataset, MSR Action3D, a depth sensor like Kinect provides an estimate of 3D joint coordinates of body skeleton, and the following features were extracted to represent MSR Action3D depth sequences -- (1) Position: 3D coordinates of the 20 joints obtained from the skeleton map. These 3D coordinates were then concatenated resulting in a 60 dimensional feature per frame; (2) Angle: normalized pair-wise angles. The normalized pair-wise angles were obtained from 18 joints of the skeleton map. The two feet joints were not included. This resulted in a 136 dimensional feature vector per frame; (3) Offset: offset of the 3D joint positions between the current and the previous frame \cite{zhu2013fusing}. These offset features were also computed using the 18 joints from the skeleton map resulting in a 54 dimensional feature per frame; (4) Velocity: histogram of the velocity components obtained from point cloud. This feature was computed using the 18 joints as in the previous cases resulting in a 162 dimensional feature per frame; (5) Pairwise joint distances: The 3D coordinates obtained from the skeleton map were used to compute pairwise joint distances with the centre of the skeleton map resulting in a 60 dimensional feature vector per frame. For the following experiments, these five different features were concatenated to result in a $583$ dimensional feature vector per frame.


\subsection{Architecture and Training}

\begin{table}
\begin{center}
    \begin{tabular}{ l | c | c }
    \toprule
    \textbf{Dataset} & \textbf{KTH} & \textbf{MSR Action3D}  \\ \hline
    Input Units & 450 & 583 \\
    Memory Cell State Units & 300 & 400 \\
	Output Units & 6 & 20 \\ \hline
    \end{tabular}\vspace{2mm}\caption{Architecture of the dRNN model used on two datasets.
    Each row shows the number of units in each component. For the sake of fair comparison, we adopt the same architecture for the dRNN models of both orders on two datasets.}\label{Tb:Arch}
\end{center}
\end{table}

The architectures of the dRNN models trained on the two datasets are shown in Table \ref{Tb:Arch}.  For the sake of fair comparison, we adopt the same architecture for the dRNN models of both orders on two datasets. We can see that the number of memory cell units is smaller than the input units on both datasets.  This can be interpreted as follows. The sequence of an action video often forms a continuous trajectory embedded in a low-dimensional manifold of the input space. Thus, a lower-dimension state space suffices to capture the dynamics of such a trajectory.

\begin{figure}[t]
\begin{center}
   \includegraphics[width=0.9\linewidth]{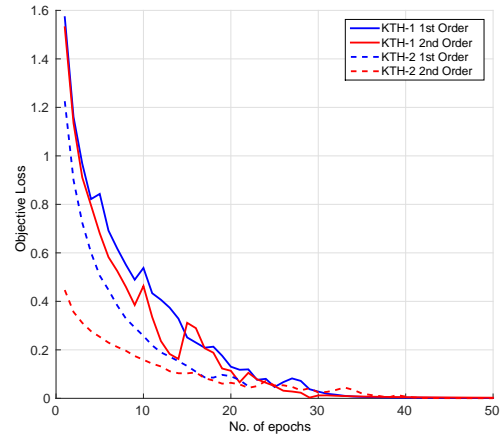}
\end{center}
   \caption{Objective loss curve over training epochs on the KTH dataset.}
\label{Fig:KTHlearnCost}
\end{figure}

We plot the learning curve for training the model on KTH dataset in Figure \ref{Fig:KTHlearnCost}.  The learning rate of BPTT algorithm is set to $0.0001$.  The figure shows that the objective loss  continuously decreases over $50$ epochs.  Usually after 40 epochs, the training of dRNN model begins to converge.

%

\begin{figure}[t]
\begin{center}
   \includegraphics[width=0.9 \linewidth]{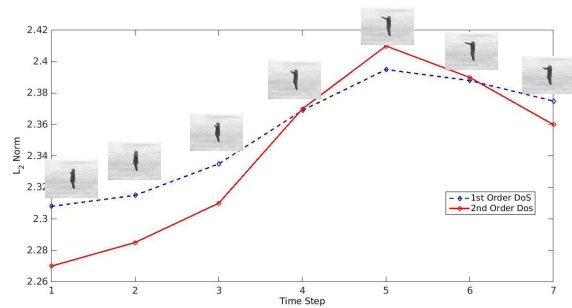}
\end{center}
   \caption{The curve of the 1st and 2nd orders of DoS over an example of sequence for the action ``boxing." Note that the local maximum of DoS corresponds to the change from punching to relaxing.  }
\label{Fig:FrameCorrespondence}
\end{figure}

\subsection{Results on KTH Dataset}

There are several different evaluation protocols used on KTH dataset in literature.  This can result in as large as $9\%$ differences in performance across different experiment protocols as reported in \cite{gao2010comparing}. For the sake of fair comparison, we follow the cross-validation protocol \cite{baccouche2011sequential}, in which we randomly select $16$ subjects to train the model, and test over the remaining $9$ subjects.  The performance is reported by the average across five such trails.

\begin{figure}[t]
\begin{center}
   \includegraphics[width=0.9\linewidth]{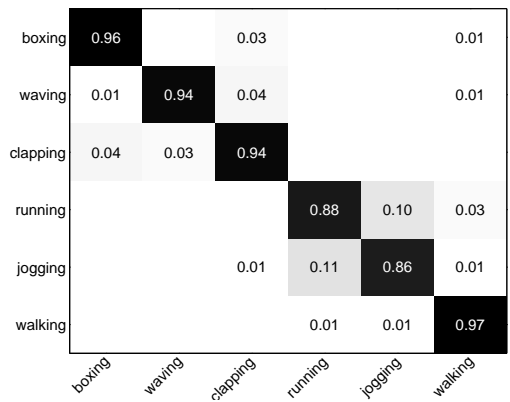}
\end{center}
   \caption{Confusion Matrix on the KTH-1 dataset obtained by the 2-Order dRNN model.}
\label{Fig:KTHconf}
\end{figure}

First, we compare the dRNN model with the conventional LSTM model in Table \ref{Tb:Table0}. Here we report the cross-validation accuracy on both KTH-1 and KTH-2 datasets. In addition,
Figure \ref{Fig:KTHconf} shows the confusion matrix obtained by the 2-order dRNN model on KTH-1 dataset.  This confusion matrix is computed by averaging over five trials in the above cross-validation protocol.
The performance of conventional LSTM has been reported in literature \cite{grushin2013robust,baccouche2011sequential}. We note that these reported accuracies often vary with different types of features.
Thus, a fair comparison between different models can only be made with the same type of input feature.

For the dRNN model, we report the accuracy with up to the $2$-order of DoS.
The table shows that with the same HOG3D feature, the proposed dRNN models outperform the conventional LSTM model, and the $2$-order dRNN yields a better accuracy than its $1$-order counterpart.
Although higher-order of DoS might improve the accuracy further, we do not report the result since it becomes trivial to simply add more orders of DoS into dRNN, and the improved performance might not compensate for the increased computational cost. For most of practical applications, the first two orders of dRNN should be competent enough.

Baccouche et al. \cite{baccouche2011sequential} reported an accuracy of $94.39\%$ and $92.17\%$ on KTH-1 and KTH-2 data sets, respectively.  But it is worth noting that they used a combination of 3DCNN and LSTM, where 3DCNN plays the crucial role in reaching such performance. Actually, 3DCNN model alone can reach an accuracy of $91.04\%$ and $89.40\%$ on KTH-1 and KTH-2 data sets as reported in \cite{baccouche2011sequential}.
On the contrary, they reported that the LSTM with Harris3D feature only achieved $87.78\%$ on KTH-2, as compared with $92.12\%$ accuracy obtained by 2-order dRNN with HOG3D feature.  In Table \ref{Tb:Table0}, under a fair comparison with the same feature, the dRNN models of both orders outperform their LSTM counterpart with the same HOG3D feature.

In Figure \ref{Fig:FrameCorrespondence}, to support our motivation of learning LSTM representations based on the dynamic change of states evolving over frames, we illustrate some example frames of ``boxing" action versus the curve of $L_2$-norm of 1-order and 2-order DoS on KTH dataset. It shows the change from ``punching" to ``relaxing" at the local maximum of DoS, showing the ability of the dRNN model to capture the salient dynamics for the action.

We also show the performance of the other non-LSTM state-of-the-art approaches in Table \ref{Tb:Table1}. Many of these compared algorithms focus on the action recognition problem, relying on the special assumptions about the spatio-temporal structure of actions. They might not be applicable to model the other type of sequences which do not satisfy these assumptions.  In contrast, the proposed dRNN model is a general-purpose model, not being tailored to specific type of action sequences.  This also makes it competent on 3D action recognition task as we will show below.



\begin{table}
\begin{center}
    \begin{tabular}{ l | l | l }
    \toprule
    \textbf{Dataset} & \textbf{Method} & \textbf{Accuracy}  \\ \hline
	\multirow{4}{*}{KTH-1} & {LSTM + HOF } \cite{grushin2013robust}& {90.7} \\
    & {LSTM + HOG3D}	 & 89.93 \\
    & {1-order dRNN  + HOG3D} & \bf{93.28} \\
	& {2-order dRNN  + HOG3D} & \bf{93.96} \\\hline
	\multirow{5}{*}{KTH-2}
    & {LSTM + Harris3D} \cite{baccouche2011sequential}  & {87.78}\\	
    & {LSTM + HOG3D} & 87.32 \\
    & {1-order dRNN + HOG3D} & \bf{91.98} \\
	& {2-order dRNN + HOG3D} & \bf{92.12} \\\hline
    \end{tabular}\vspace{2mm}\caption{Cross-validation accuracy over five trails obtained by the proposed dRNN model in comparison with the conventional LSTM model on KTH-1 and KTH-2 data sets.}\label{Tb:Table0}
\end{center}
\end{table}

\begin{table}
\begin{center}
    \begin{tabular}{ l | l | l }
    \toprule
    \textbf{Dataset} & \textbf{Method} & \textbf{Accuracy}  \\ \hline
	\multirow{3}{*}{KTH-1}
    & Rodriguez et al. \cite{rodriguez2008action}  & 81.50 \\	
    & Jhuang et al.\cite{jhuang2007biologically} & {91.70} \\
	& Schindler et al. \cite{schindler2008action} & 92.70 \\
    & 3DCNN \cite{baccouche2011sequential} & 91.04 \\ \hline
	\multirow{5}{*}{KTH-2}
	& {Ji et al.} \cite{ji20133d} & {90.20} \\
	& {Taylor et al.} \cite{taylor2010convolutional} & {90.0} \\
	& {Laptev et al.} \cite{laptev2008learning} & 91.80 \\
	& {Dollar et al.} \cite{dollar2005behavior} & 81.20 \\
    & {3DCNN} \cite{baccouche2011sequential} & {89.40} \\\hline
    \end{tabular}\vspace{2mm}\caption{Cross-validation accuracy over five trials obtained by the other compared algorithms on KTH-1 and KTH-2 datasets.}\label{Tb:Table1}
\end{center}
\end{table}



\begin{table}
\begin{center}
    \begin{tabular}{ l | l }
    \toprule
    \textbf{Method} & \textbf{Accuracy}  \\ \hline
    Actionlet Ensemble \cite{wang2012mining} & 88.20 \\ \hline
    HON4D \cite{oreifej2013hon4d} & 88.89 \\ \hline
    DCSF \cite{xia2013spatio} & 89.3 \\ \hline
    Lie Group \cite{vemulapalli2014human} & 89.48 \\ \hline
    LSTM & 87.78 \\\hline
    {1-order dRNN} & \textbf{91.40} \\ \hline	
    {2-order dRNN} & \textbf{92.03} \\ \hline
    \end{tabular}\vspace{2mm}\caption{Comparison of the dRNN model with the other algorithms on MSR Action3D dataset.}\label{Tb:Table2}
\end{center}
\end{table}

\subsection{Results on MSR Action3D Dataset}

\begin{figure}[t]
\begin{center}
   \includegraphics[width=0.9\linewidth]{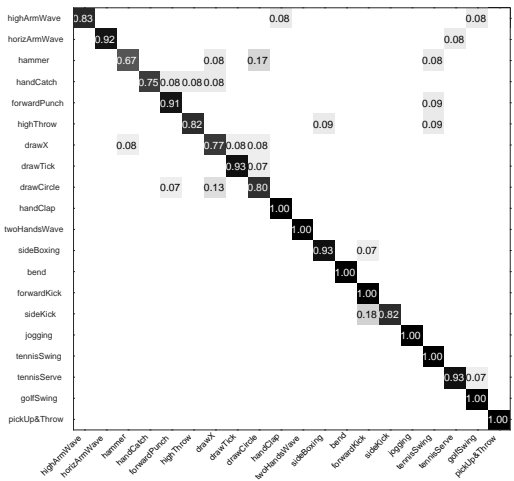}
\end{center}
   \caption{Confusion Matrix on the MSR Action3D dataset by the 2-Order dRNN model}\label{fig:confMSR}
\end{figure}

Table \ref{Tb:Table2} compares the results on MSR Action3D dataset, and Figure \ref{fig:confMSR} shows the confusion matrix by the 2-order dRNN model.
The results are obtained by following exactly the same experimental setting in \cite{wang2012mining}, in which half of actor subjects are used for training and the rest are used for testing.   This is in contrast to another evaluation protocol in literature \cite{li2010action} that splits across $20$ action classes into three subsets and performs the evaluation within each individual subset. The evaluation protocol we adopt is more challenging because it is evaluated over all $20$ action classes with no common subjects in training and test sets.

From the results, the dRNN models of both orders outperform the conventional LSTM algorithm with the same feature.  Also, both dRNN models perform competitively as compared with the other algorithms.
We notice that the Super Normal Vector (SNV) model \cite{yang2014super} has reported an accuracy of $93.09\%$ on MSR Action3D dataset.  However, this model is based on a special assumption about the 3D geometric structure of the surfaces of depth image sequences.  Thus, this approach is a very special model for solving 3D action recognition problem.  This is contrary to dRNN as a general model without any specific assumptions on the dynamic structure of the video sequences.

In brief, through the experiments on both 2D and 3D human action datasets, we show the competitive performance of dRNN compared with both LSTM and non-LSTM models. This demonstrates its wide applicability in representing and modeling the dynamics of both 2D and 3D action sequences, irrespective of any assumptions on the structure of video sequences.

\section{Conclusion and Future Work}\label{sec:clu}

In this paper, we present a new family of differential Recurrent Neural Networks (dRNNs) that extend Long Short-Term Memory (LSTM) structure by modeling the dynamics of states evolving over time.  The new structure is better at learning the salient spatio-temporal structure.  Its gate units are controlled by the different orders of derivatives of states, making the dRNN model more adequate for the representation of the long short-term dynamics of actions. Experiment results on both 2D and 3D human action datasets demonstrate the dRNN model outperforms the conventional LSTM model.  Even in comparison with the other state-of-the-art approaches based on strong assumptions about the motion structure of actions being studied, the general-purpose dRNN model still demonstrates much competitive performance on both 2D and 3D datasets. In the future work, we will test dRNN in combination with more sophisticated input feature sequences to explore the specific motion structure of actions.


{\small
\bibliographystyle{ieee}
\bibliography{egbib}
}

\end{document}